\documentclass[letterpaper]{article} 
\usepackage{aaai2026}  
\usepackage{times}  
\usepackage{helvet}  
\usepackage{courier}  
\usepackage[hyphens]{url}  
\usepackage{graphicx} 
\usepackage{amsfonts} 
\usepackage{natbib}  
\usepackage{caption} 
\usepackage{algorithm}
\usepackage{algorithmic}

\usepackage{amsmath}
\usepackage{amssymb}
\usepackage[table,xcdraw,dvipsnames]{xcolor}
\usepackage{booktabs}
\usepackage{multirow}
\usepackage{subfig}
\usepackage{makecell}

\usepackage[table]{xcolor}
\usepackage{pgfmath}
\usepackage{booktabs}
\usepackage{array}

\newcommand{\heatcellA}[2]{%
  \begingroup
  \pgfmathsetmacro{\value}{#1}
  \pgfmathsetmacro{\min}{72}   
  \pgfmathsetmacro{\max}{81}   
  \pgfmathsetmacro{\ratio}{(\value - \min)/(\max - \min)}
  \pgfmathsetmacro{\r}{round(255 - 25 * \ratio)}
  \pgfmathsetmacro{\g}{round(255 - 10 * \ratio)}
  \pgfmathsetmacro{\b}{round(255 - 20 * \ratio)}

  \xdef\cellcol{rgb,255:red,\r;green,\g;blue,\b}%
  \cellcolor{\cellcol}%
  \makebox[3em][c]{\strut
    \ifnum#2=1
      \textbf{#1}%
    \else
      #1%
    \fi
  }%
  \endgroup
}

\newcommand{\heatcellB}[2]{%
  \begingroup
  \pgfmathsetmacro{\value}{#1}
  \pgfmathsetmacro{\min}{80}   
  \pgfmathsetmacro{\max}{96}   
  \pgfmathsetmacro{\ratio}{(\value - \min)/(\max - \min)}
  \pgfmathsetmacro{\r}{round(255 - 25 * \ratio)}
  \pgfmathsetmacro{\g}{round(255 - 10 * \ratio)}
  \pgfmathsetmacro{\b}{round(255 - 20 * \ratio)}

  \xdef\cellcol{rgb,255:red,\r;green,\g;blue,\b}%
  \cellcolor{\cellcol}%
  \makebox[3em][c]{\strut
    \ifnum#2=1
      \textbf{#1}%
    \else
      #1%
    \fi
  }%
  \endgroup
}

\usepackage{newfloat}
\usepackage{listings}
\DeclareCaptionStyle{ruled}{labelfont=normalfont,labelsep=colon,strut=off} 
\floatstyle{ruled}
\newfloat{listing}{tb}{lst}{}
\floatname{listing}{Listing}
\title{HyPCV-Former: Hyperbolic Spatio-Temporal Transformer for 3D Point Cloud Video Anomaly Detection}
\author{
    Jiaping Cao\textsuperscript{\rm 1},  Kangkang Zhou\textsuperscript{\rm 2}, Juan Du\textsuperscript{\rm 1}\textsuperscript{\rm 3}\thanks{Corresponding author: juandu@hkust-gz.edu.cn}
}
\affiliations{
    \textsuperscript{\rm 1}The Hong Kong University of Science and Technology (Guangzhou)\\
    \textsuperscript{\rm 2}University of Chinese Academy of Sciences\\
    \textsuperscript{\rm 3}The Hong Kong University of Science and Technology

}

\usepackage{bibentry}

\begin{document}
\makeatletter
\def\copyright@text{}
\makeatother
\maketitle

\begin{abstract}
Video anomaly detection is a fundamental task in video surveillance, with broad applications in public safety and intelligent monitoring systems. Although previous methods leverage Euclidean representations in RGB or depth domains, such embeddings are inherently limited in capturing hierarchical event structures and spatio-temporal continuity. To address these limitations, we propose HyPCV-Former, a novel hyperbolic spatio-temporal transformer for anomaly detection in 3D point cloud videos. Our approach first extracts per-frame spatial features from point cloud sequences via point cloud extractor, and then embeds them into Lorentzian hyperbolic space, which better captures the latent hierarchical structure of events. To model temporal dynamics, we introduce a hyperbolic multi-head self-attention (HMHA) mechanism that leverages Lorentzian inner products and curvature-aware softmax to learn temporal dependencies under non-Euclidean geometry. Our method performs all feature transformations and anomaly scoring directly within full Lorentzian space rather than via tangent space approximation. Extensive experiments demonstrate that HyPCV-Former achieves state-of-the-art performance across multiple anomaly categories, with a 7\% improvement on the TIMo dataset and a 5.6\% gain on the DAD dataset compared to benchmarks. The code will be released upon paper acceptance.
\end{abstract}


\section{Introduction}

\begin{figure}[ht]
  \centering
  \includegraphics[width=\linewidth]{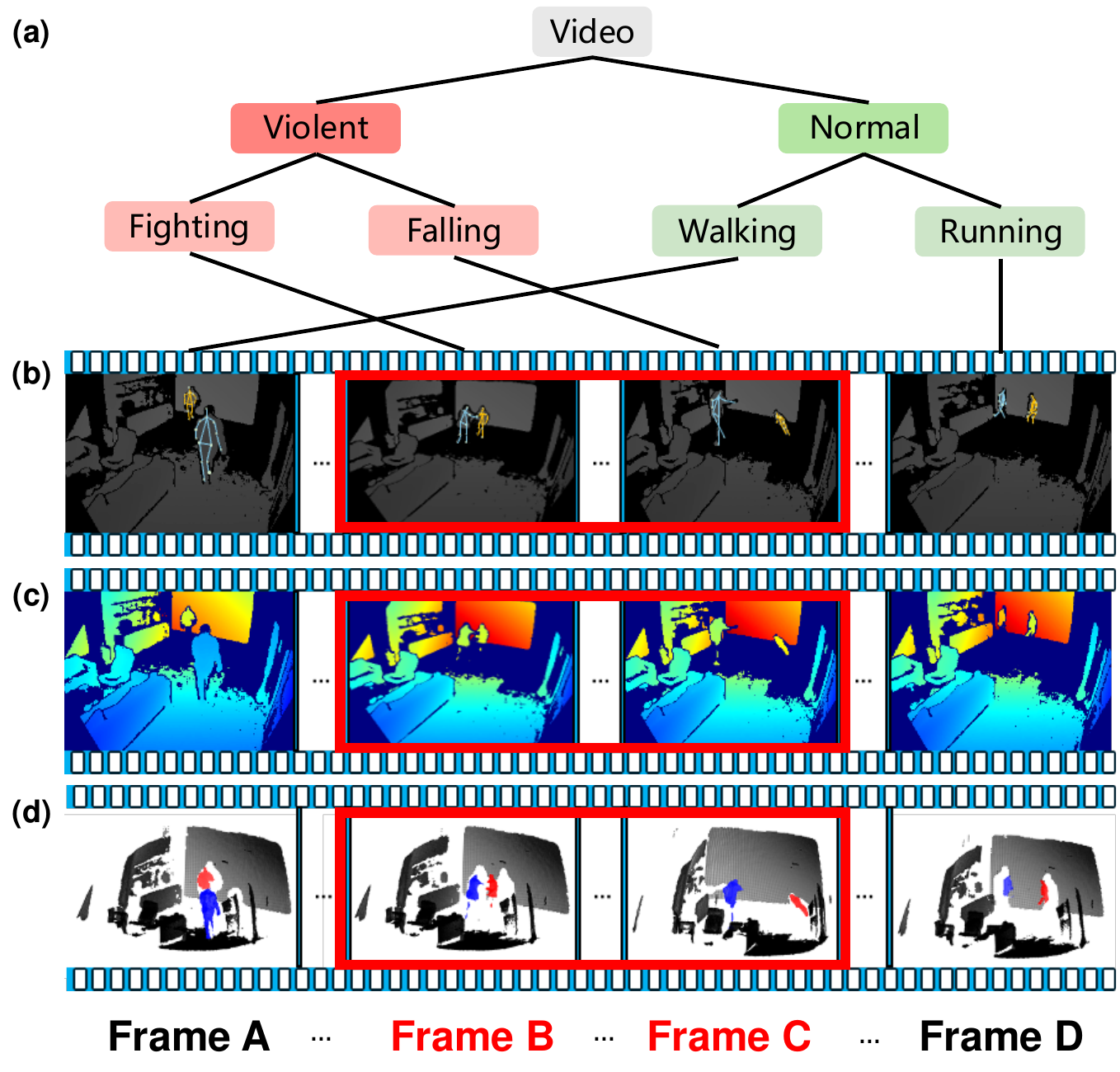}
  \caption{Illustration of VAD and 3D data acquisition. (a) Hierarchical diagram in VAD. (b-d) Visualization of video frames using (b) human pose estimation, (c) depth images, and (d) 3D point clouds. Frames A and D are normal; Frames B and C are anomalous.}
  \label{fig:1}
\end{figure}
Video anomaly detection (VAD), often referred to as video violence detection, is a fundamental task in video surveillance that aims to identify abnormal events deviating from expected patterns~\cite{leng2024beyond}. Unlike anomalies typically found in images, violent or anomalous events in videos are challenging to detect from single frames, as these frames often lack obvious geometric defects or color irregularities. Therefore, VAD requires analyzing a sequence of frames that collectively illustrate hierarchical structures—comprising frames before, during, and after an anomaly—as depicted in Figure~\ref{fig:1}(a). Hyperbolic space is particularly effective for representing hierarchical or tree structures due to its exponential relationship between node quantity and tree depth, contrasting with the polynomial relationship characteristic of Euclidean space. Existing studies have applied hyperbolic representation learning to VAD tasks~\cite{leng2024beyond,peng2023learning}. However, both HyperVD~\cite{peng2023learning} and DSRL~\cite{leng2024beyond} define operations in tangent space rather than directly within hyperbolic space, compromising computational accuracy.


Most existing methods for VAD rely on RGB images~\cite{karim2024real,yang2024text}, which provide rich semantic information. However, anomaly detection based on RGB images is often sensitive to lighting conditions and lacks aggregated defect features. To leverage accurate 3D spatial information instead of conventional 2D data, some approaches utilize human pose estimation~\cite{zhang2025esmformer,zhang2024deep} or range images~\cite{schneider2022unsupervised,kopuklu2021driver}, as illustrated in Figure~\ref{fig:1}(b) and (c). Nevertheless, pose estimation methods typically reconstruct 3D coordinates indirectly from 2D images, and range images used in anomaly detection tasks are computationally intensive and unsuitable for analyzing complex spatial structures~\cite{bergmann2021mvtec}. In contrast, 3D point clouds are inherently unstructured and unordered, consisting of discrete points distributed across object surfaces. 

3D point cloud videos, also referred to as 3D point cloud streams, consist of sequential frames of 3D point clouds~\cite{zhang2021cloudlstm}. To our knowledge, anomaly detection within 3D point cloud videos has not yet been extensively explored. Notably,~\cite{he2024point} directly utilize 3D point clouds to analyze human dynamics and detect anomalies, yet they learn point representations solely in Euclidean space. Given the proven advantages of 3D point clouds~\cite{xie2020linking,uy2019revisiting}, along with the effectiveness of hyperbolic representation learning~\cite{peng2020mix,long2020searching,hong2023curved}, we propose exploring this data modality within hyperbolic space for VAD, as depicted in Figure~\ref{fig:1}(d).


To address aforementioned drawbacks and challenges, we propose a novel 3D point cloud video anomaly detection recipe based on hyperbolic spatio-temporal transformer, abbreviating to HyPCV-Former. We use 3D point clouds modality to effectively capture geometric features~\cite{he2024point} and introduce hyperbolic space to differentiate between normal and anomalous events—particularly ambiguous violence in surveillance videos~\cite{leng2024beyond,peng2023learning}. In order to better learn the temporal dependency within frame sequences, we design the hyperbolic multi-head self-attention (HMHA) mechanism. Lorentzian intrinsic distance is used as the anomaly score, ensuring that all computations are performed entirely within hyperbolic space. Our key contributions can be summarized as follows:
\begin{itemize}
\item We propose a hyperbolic spatio-temporal transformer for 3D point cloud videos, which amplifies the separation between normal and abnormal instances to enhance anomaly discrimination.

\item We introduce an HMHA mechanism operating entirely in Lorentzian space to model frame-level dynamics and enhance spatio-temporal representations for anomaly prediction.

\item To the best of our knowledge, HyPCV-Former is the first to leverage hyperbolic geometry for anomaly detection in 3D point cloud videos, achieving state-of-the-art performance on violent event detection.

\end{itemize}

\section{Related Work}

\textbf{3D Point Cloud Analysis} 3D Point cloud data has received a lot of attention for its superior accuracy and robustness in a variety of adverse situations~\cite{xiao2023unsupervised}. Some previous works firstly transfer point clouds into octrees~\cite{hornung2013octomap} or hashed voxel lists~\cite{niessner2013real} due to the unordered property. Others use point-based deep architecture to learn individual point representation through stacking several multilayer perceptron (MLP), such as PointNet~\cite{qi2017pointnet} and PointNet++~\cite{qi2017pointnet++}. In addition, there are some existing works that regard point clouds as graphs in Euclidean space to capture dependencies between adjacent points. For example, DGCNN~\cite{wang2019dynamic} dynamically constructs a graph at each layer, where the edges are defined by the k-nearest neighbors in feature space. PointMLP~\cite{ma2022rethinking} discards sophisticated local geometric extractors in favor of a lightweight geometric affine module and a deeper MLP design. This approach efficiently captures and fuses local geometry, achieving leading performance on multiple datasets. Thus, we introduce 3D point clouds to obtain geometric characteristics of each frame, which can be better projected into features containing spatial information related to anomalies and protecting personal information. 

\textbf{Video Anomaly Detection} Due to the rarity and unpredictability of anomalous events, compiling exhaustive labeled datasets encompassing all potential anomalies is generally infeasible~\cite{schneider2022unsupervised}. Consequently, unsupervised methods are particularly advantageous, as they primarily seek to model normal spatio-temporal patterns without presupposing the nature of anomalies. These methods detect outliers by identifying substantial deviations from the learned manifold of standard motion and appearance. Unsupervised approaches in this domain are primarily categorized into reconstruction-based and prediction-based methods~\cite{liu2024generalized}. However, these recipes are predominantly applied to RGB video, where the lack of precise coordinate information and potential disclosure of personal details remain notable concerns~\cite{he2024point}.

To address these issues, a range of 3D-based strategies for anomaly detection have been explored, encompassing human pose estimation, depth imaging, and point cloud video streams. Specifically, Zhang \textit{et al.}~\cite{zhang2025esmformer, zhang2024deep} employ human pose estimation approaches to extract 3D human information from frame sequences. Besides, Schneider \textit{et al.}~\cite{schneider2022unsupervised} employ Time-of-Flight (ToF) depth images to capture 3D information for unsupervised video anomaly detection, highlighting the advantages of depth data in robust geometric representation and reduced privacy risk. He \textit{et al.}~\cite{he2024point} leverage point cloud video sequences to obtain more precise 3D spatial information for video anomaly detection. By converting each depth frame into a 3D point cloud representation and applying a specialized reconstruction-based autoencoder, their approach captures fine-grained geometric and motion details while maintaining privacy. Therefore, our framework follows the prediction-based anomaly detection paradigm to avoid preserving anomalous pattern.

\textbf{Hyperbolic Learning} In representation learning, hyperbolic geometry has been extensively studied for its ability to model complex non-Euclidean data, demonstrating enhanced representational capacity and generalization when dealing with hierarchical structures~\cite{montanaro2022rethinking}. Leveraging hyperbolic space, numerous neural network architectures have been devised to exploit the geometric advantages it offers~\cite{ganea2018hyperbolic,chami2019hyperbolic,liu2019hyperbolic}. Hyperbolic geometry has found wide-ranging applications in computer vision~\cite{mettes2024hyperbolic,khrulkov2020hyperbolic,ermolov2022hyperbolic}, recommendation~\cite{shimizu2024fashion, tan2022enhancing}, and graph learning~\cite{dai2021hyperbolic, sun2021hyperbolic, du2024efficient}. Leng \textit{et al.}~\cite{leng2024beyond} propose a hyperbolic neural network–based framework for video violence detection, leveraging hyperbolic space to capture hierarchical structures more effectively. However, they also highlight that relying too frequently on hyperbolic operations risks destabilizing the training process, underscoring the need for strategies that maintain stable representation learning.

Recently, there have also been efforts to adapt Transformers and other neural architectures to hyperbolic or mixed-curvature spaces. Gulcehre \textit{et al.}~\cite{gulcehre2018hyperbolic} propose hyperbolic attention networks, which reinterpret the standard soft-attention mechanism in hyperbolic geometry to better encode hierarchical data and empirically verify the resulting improvements. Chen \textit{et al.}~\cite{chen2021fully} extend this direction by formulating a fully hyperbolic network in the Lorentz model, avoiding reliance on tangent-space transformations and thereby enhancing both representation ability and computational stability. Shimizu \textit{et al.}~\cite{shimizu2020hyperbolic} further refine hyperbolic network components under the HNN++ framework, introducing more parameter-efficient implementations of hyperbolic multinomial logistic regression and convolutional layers. Finally, Cho \textit{et al.}~\cite{cho2023curve} target mixed-curvature Transformers, demonstrating that by end-to-end learning of curvature parameters, the architecture can flexibly adapt to complex relational structures and improve performance on various graph-centered tasks.

\section{Preliminaries}

\begin{figure*}[t]
  \centering
  \includegraphics[width=\linewidth]{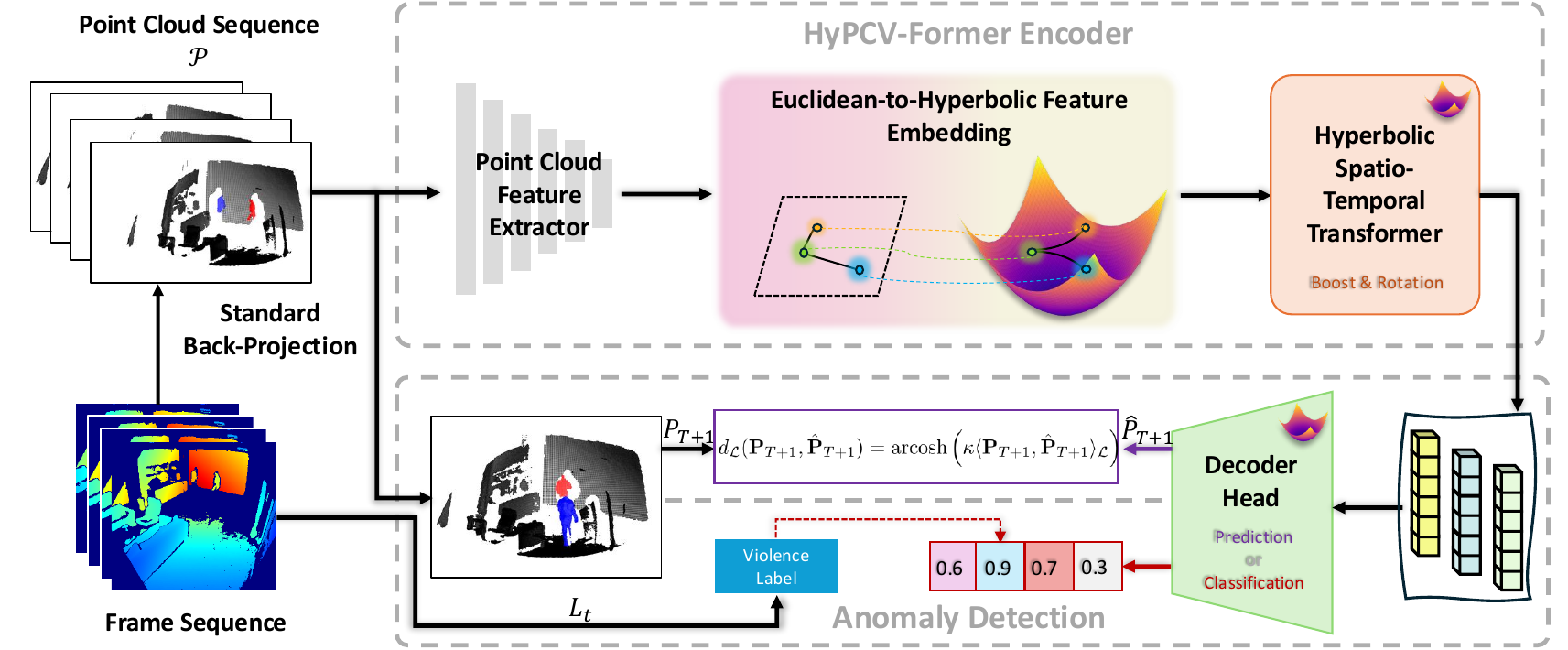}
  \caption{The pipeline of HyPCV-Former.}
  \label{fig:all}
\end{figure*}

\subsection{Problem Definition}

Point cloud video anomaly detection aims to identify anomalous events in spatial and temporal dimensions within a sequence of 3D point cloud frames. Formally, a point cloud video is represented as a sequence $\mathcal{P} = \{\mathbf{P}_1, \mathbf{P}_2, \dots, \mathbf{P}_T\}$, where $t$-th frame $\mathbf{P}_t \in \mathbb{R}^{N \times 3}$ consists of $N$ points, with each point described by its 3D coordinates $(x, y, z)$. Each frame $\mathbf{P}_t$ has its label $L_t$ that can supervise the training of anomaly score. In our study, we regard VAD as a prediction or a classification task according to different benchmarks.


At testing time, the model evaluates a point cloud sequence $\mathcal{P}$ and assigns an anomaly score $s_{T+1}$ to the target frame $\mathbf{P}_{T+1}$ based on its deviation from learned spatio-temporal patterns in the prediction setting:
\begin{equation}
s_{T+1} = \mathcal{D}\left(\mathbf{P}_{T+1}, f_\theta(\mathcal{P})\right),
\end{equation}
where $\mathcal{D}(\cdot, \cdot)$ denotes a prediction-based distance. Frames with high anomaly scores are flagged as abnormal, indicating significant deviations from expected behavior. In classification setting, the objective is to learn a model $f_\theta$ that assigns an anomaly score to the target frame, reflecting the likelihood of abnormal behavior.



\subsection{Lorentz Model for Hyperbolic Geometry}

Hyperbolic geometry effectively models hierarchical structures, and the Lorentz model is particularly favored for its efficiency, numerical stability, and well-defined operations~\cite{chen2021fully,peng2023learning}. 

\subsubsection{Lorentz Model}
Formally, the Lorentz model, also known as the hyperboloid model, is represented as an n-dimensional Riemannian manifold $\mathbb{L}^n_{\kappa} = (\mathcal{L}^n, g^\kappa_\textbf{x})$, where $\kappa < 0$ indicates a negative constant curvature. The set $\mathcal{L}^n$ describes an upper hyperboloid sheet in an $(n+1)$-dimensional Minkowski space, given by:

\begin{equation}
  \mathcal{L}^n = \{\textbf{x} \in \mathbb{R}^{n+1} : \langle \textbf{x},\textbf{x} \rangle_{\mathcal{L}} = 1/\kappa, \textbf{x}_0 > 0\},
\end{equation}
where $\langle \cdot, \cdot \rangle_{\mathcal{L}}$ is the Lorentzian scalar product defined as:

\begin{equation}
    \langle \textbf{x},\textbf{y} \rangle_{\mathcal{L}} = -x_0y_0 + \sum_{i=1}^{n} x_iy_i.
\end{equation}

The Lorentzian scalar product distinguishes the temporal axis $(x_0)$ from the spatial axes $(x_i, i \geq 1)$, borrowing terminology from special relativity.  

\subsubsection{Tangent Space} 
At any point $\textbf{x} \in \mathbb{L}_\kappa^n$, the tangent space $\mathcal{T}_\textbf{x}\mathbb{L}_\kappa^n$ represents a local linear approximation to the hyperboloid and is defined by:
\begin{equation}
    \mathcal{T}_\textbf{x}\mathbb{L}_\kappa^n = \{\textbf{y} \in \mathbb{R}^{n+1} : \langle \textbf{y},\textbf{x} \rangle_{\mathcal{L}} = 0\}.
\end{equation}

This tangent space is essentially a Euclidean subspace embedded in $\mathbb{R}^{n+1}$ and is crucial for performing optimization and mapping operations in hyperbolic neural networks~\cite{yang2024hypformer}.

\subsubsection{Exponential and Logarithmic Maps} 
Operations such as neural network training in hyperbolic spaces require seamless transitions between hyperbolic manifolds and their tangent spaces. This transition is achieved via exponential and logarithmic maps, essential mathematical tools defined in hyperbolic geometry~\cite{leng2024beyond}. The exponential map $\text{exp}_\textbf{x}^\kappa$ transfers a vector from the tangent space at point $x$ to the hyperbolic manifold, mathematically defined as:
\begin{equation}
\label{eq5}
    \text{exp}_\textbf{x}^\kappa(\textbf{v}) = \cosh(\alpha)\textbf{x} + \sinh(\alpha)\frac{\textbf{v}}{\alpha}, \quad \alpha = \sqrt{-\kappa\langle \textbf{v},\textbf{v} \rangle_{\mathcal{L}}}.
\end{equation}

Conversely, the logarithmic map $\text{log}_\textbf{x}^\kappa$ projects points from the manifold back onto the tangent space:
\begin{equation}
    \text{log}_\textbf{x}^\kappa(\textbf{y}) = \frac{\text{arcosh}(\beta)}{\sqrt{\beta^2 - 1}}(\textbf{y} - \beta \textbf{x}), \quad \beta = \kappa\langle \textbf{x},\textbf{y} \rangle_{\mathcal{L}}.
\end{equation}

These mappings enable efficient geometric manipulations necessary for representation learning and model optimization within hyperbolic spaces, particularly in tasks involving hierarchical data structures such as anomaly detection in 3D point clouds~\cite{li2025hyperbolic}.

\section{Method}
In this section, we present the overall framework of HyPCV-Former for anomaly detection in 3D point cloud videos, as depicted in Figure~\ref{fig:all}. Our approach consists of two stages: (1) HyPCV-Former encoder, (2) anomaly detection. There are two operations in the encoder, including hyperbolic representation learning and hyperbolic spatio-temporal transformation. Based on the learned representations, an anomaly score is computed for the target frame using a task-specific criterion defined in hyperbolic space.

\subsection{Hyperbolic Representation}

\subsubsection{Per-frame Point Cloud Feature Extraction} 

To extract frame-wise features from each point cloud $\mathbf{P}_t \in \mathbb{R}^{N \times 3}$, we employ PointNet~\cite{qi2017pointnet} for its efficiency and permutation invariance. Other backbones such as PointMLP~\cite{ma2022rethinking} and DGCNN~\cite{wang2019dynamic} are also compatible with our framework. Each frame is encoded into a $D$-dimensional Euclidean feature vector: 

\begin{equation}
    \mathbf{x}_t=\text{PointNet}(\mathbf{P}_t)\in\mathbb{R}^{D}.
\end{equation}

Thus, the video sequence is transformed into a Euclidean tensor $\mathbf{X}= [\mathbf{x}_1,\mathbf{x}_2,\dots,\mathbf{x}_T]\in\mathbb{R}^{B\times T\times D}$, where $B$ is the batch size, $T$ is the number of frames, and $D$ is the dimensionality of features per-frame.

\subsubsection{Euclidean-to-Hyperbolic Feature Embedding} The extracted Euclidean features inherently fail to encode hierarchical or complex structured information. To map Euclidean features onto hyperbolic manifold $\mathbb{L}_{\kappa}^{D}$, we first associate each Euclidean feature vector $\mathbf{x}_t\in\mathbb{R}^{D}$ with a vector on the tangent space $\mathcal{T}_{\mathbf{o}}\mathbb{L}_{\kappa}^{D}$ at the manifold origin $\mathbf{o} = (\sqrt{-1/\kappa},0,\dots,0)$:
\begin{equation}
    \mathbf{x}_t^{(0)} = [0, \mathbf{x}_t]\in \mathcal{T}_{\mathbf{o}}\mathbb{L}_{\kappa}^{D}.
\end{equation}

Then, we use the exponential map $\mathbf{x}_t^{\mathbb{L}} = \exp_{\mathbf{o}}^{\kappa}\left(\mathbf{x}_t^{(0)}\right)$ at the origin $\mathbf{o}$ to project $\mathbf{x}_t^{(0)}$ onto the hyperbolic manifold $\mathbb{L}_{\kappa}^{D}$. Consequently, the Euclidean features from the entire frame sequence are transformed into a hyperbolic representation $\mathbf{X}^{\mathbb{L}} = [\mathbf{x}_1^{\mathbb{L}},\mathbf{x}_2^{\mathbb{L}},\dots,\mathbf{x}_T^{\mathbb{L}}] \in \mathbb{L}_\kappa^{B\times T\times (D+1)}$.

\subsection{Hyperbolic Spatio-Temporal Transformation}

\begin{figure}[ht]
  \centering
  \includegraphics[width=0.9\linewidth]{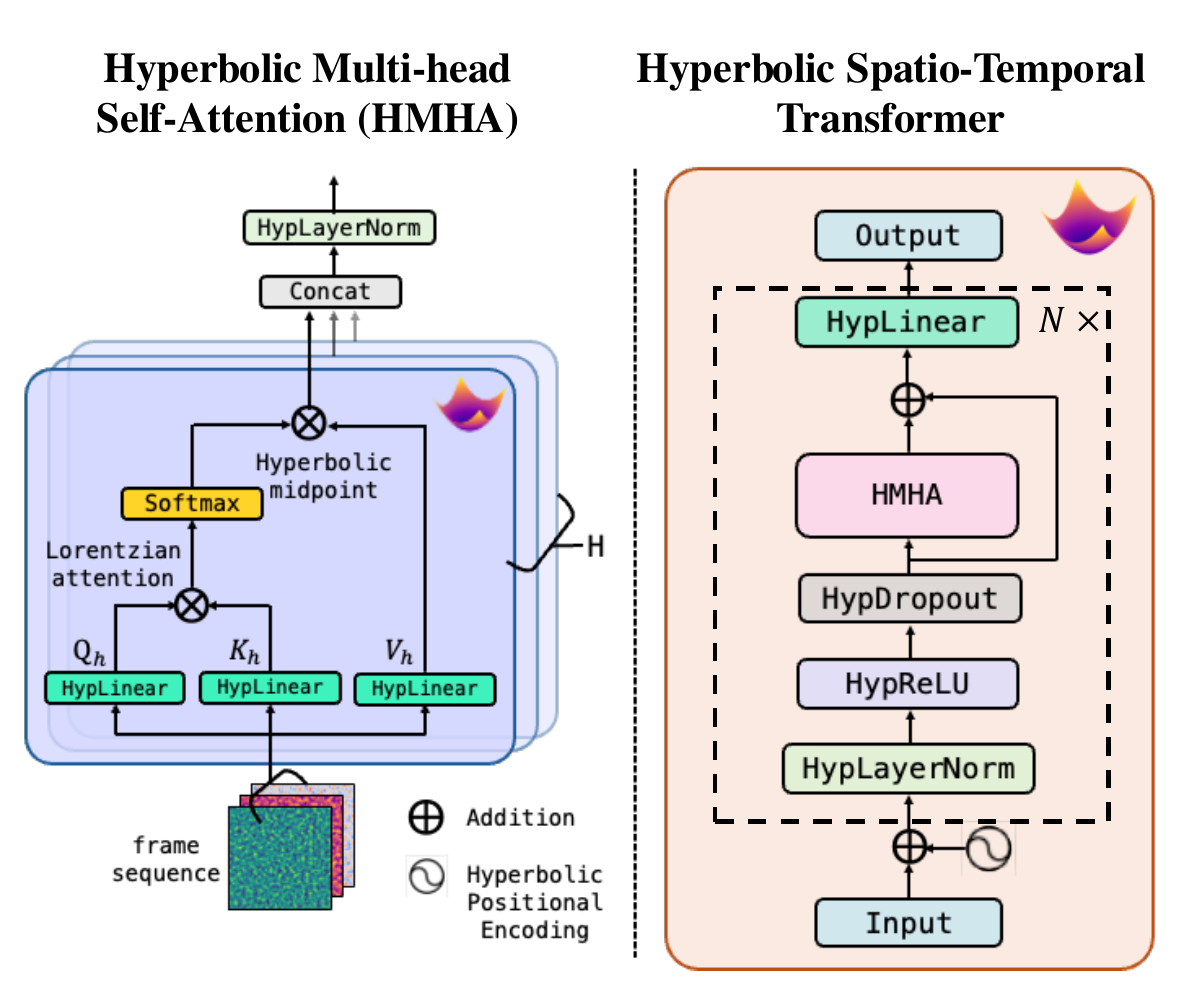}
  \caption{The details of Hyperbolic Multi-head Self-Attention and Hyperbolic Spatio-Temporal Transformer.}
  \label{fig:multihead}
\end{figure}

We propose a Hyperbolic Spatio-Temporal Transformer to effectively capture complex spatio-temporal dependencies in hyperbolic embeddings, as shown in Figure~\ref{fig:multihead}. Specifically, given a Lorentzian representation $\mathbf{X}^{\mathbb{L}} \in \mathbb{L}_\kappa^{B \times T \times (D+1)}$, our model builds on two core modules, including hyperbolic transformation with curvatures (HTC) and hyperbolic readjustment and refinement with curvatures (HRC)~\cite{yang2024hypformer}. Detailed formulations of HTC and HRC are provided in Appendix~\ref{A1}. 

\subsubsection{Hyperbolic Positional Encoding} To explicitly incorporate sequential information, we introduce learnable hyperbolic positional embeddings $\mathbf{p}_t:= \mathrm{HTC}(\mathbf{x}^{\mathbb{L}}_{t})\in\mathbb{L}_\kappa^{D+1}$. Given the original hyperbolic embeddings $\mathbf{x}^{\mathbb{L}}_t$, the positional encoding operation is defined fully within hyperbolic space as~\cite{yang2024hypformer,law2019lorentzian,vaswani2017attention}:
\begin{equation}
    \mathbf{x}^{\prime}_{t} = \frac{\mathbf{x}^{\mathbb{L}}_{t} + \epsilon \cdot \mathbf{p}_{t}}{\sqrt{|\kappa|\|\mathbf{x}^{\mathbb{L}}_{t} + \epsilon \cdot \mathbf{p}_{t}\|_\mathcal{L}}},
\end{equation}
where $\epsilon$ is a scaling hyperparameter which is set to 1 in the experiment, and the Lorentzian norm $\|\cdot\|_\mathcal{L}$ ensures that the resulting embedding remains valid on the hyperbolic manifold. Such positional encoding explicitly embeds temporal information into the representations $\mathbf{X}^{\prime} = [\mathbf{x}_1^{\prime},\mathbf{x}_2^{\prime},\dots,\mathbf{x}_T^{\prime}] \in \mathbb{L}_\kappa^{B\times T\times (D+1)}$.

\subsubsection{Hyperbolic Multi-head Self-Attention} The architecture of the HMHA is shown in Figure~\ref{fig:multihead}. For each attention head $h$, we first compute query $\mathbf{Q}_h$, key $\mathbf{K}_h$, and value $\mathbf{V}_h$ through hyperbolic linear transformations defined as follows:
\begin{equation}
\mathbf{Q}_{h}=\text{HTC}(\mathbf{X}^{\prime};\mathbf{W^\mathbf{Q}}\mathbf{Q}_h)
\end{equation}
\begin{equation}
\mathbf{K}_{h}=\text{HTC}(\mathbf{X}^{\prime};\mathbf{W^\mathbf{K}}\mathbf{K}_h)
\end{equation}
\begin{equation}
\mathbf{V}_{h}=\text{HTC}(\mathbf{X}^{\prime};\mathbf{W^\mathbf{V}}\mathbf{V}_h),
\end{equation}
where $\mathbf{Q}_{h},\mathbf{K}_{h},\mathbf{V}_{h}\in \mathbb{L}_K^{B\times T\times D_h}$, $D_h=(D+1)/H$, and $H$ represents the number of heads.


Attention is computed using negative Lorentzian distances, where the full attention matrix is constructed via Lorentzian inner products followed by softmax normalization, as defined in Equation~(\ref{eq14}). This enables effective modeling of complex spatio-temporal dependencies in sequential data.
\begin{equation}
\label{eq14}
A_{h}=\text{softmax}\left(\frac{2+2\langle\mathbf{Q}_{h},\mathbf{K}_{h}\rangle_\mathcal{L}}{\sqrt{D_h}}+b_h\right),
\end{equation}
where $A_{h}$ is the attention weight for head $h$, and $b_h$ is a learnable scalar bias.

We aggregate these attention-weighted values using the Lorentzian midpoint~\cite{law2019lorentzian}:
\begin{equation}
    A_{h} \odot^{\kappa} \mathbf{V}_{h} := 
    \frac{  A_{h} \mathbf{V}_{h} }
    { \sqrt{ |\kappa| } \left\|  A_{h} \mathbf{V}_{h} \right\|_{\mathcal{L}} }.
\end{equation}
where $A_{h} \odot^{\kappa} \mathbf{V}_{h}$ is denoted as $\mathbf{O}_{h}$. Then, concatenating the outputs from all heads, we obtain the final attention output embeddings:
\begin{equation}
    \mathbf{O}=[\mathbf{O}_1\|\mathbf{O}_2\|\dots\|\mathbf{O}_H]\in\mathbb{L}_\kappa^{B\times T\times(D+1)}.
\end{equation}

\subsubsection{Hyperbolic Non-Linear Operation} Layer normalization, activation, and dropout are essential components of the Hyperbolic Spatio-Temporal Transformer. We implement all these non-linear operations using the HRC module, which enables curvature-aware refinement within the Lorentzian manifold. Specifically, the refining function $f_r(\cdot)$ encapsulates hyperbolic layer normalization $f_{\text{LayerNorm}}(\cdot)$, hyperbolic activation $f_{\sigma}(\cdot)$, and hyperbolic dropout $f_{\text{Dropout}}(\cdot)$, as formally defined as follows:
\begin{equation}
\label{eq18}
\begin{aligned}
\text{HypLayerNorm}(\mathbf{O}) &= \text{HRC}(\mathbf{O}, f_{\text{LayerNorm}}), \\
\text{HypReLU}(\mathbf{O}) &= \text{HRC}(\mathbf{O}, f_{\sigma}), \\
\text{HypDropout}(\mathbf{O}) &= \text{HRC}(\mathbf{O}, f_{\text{Dropout}}).
\end{aligned}
\end{equation}

The output of HRC is a refined hyperbolic embedding $\mathbf{\hat{X}}=\text{HRC}(\mathbf{O};f_r,\kappa_1,\kappa_2)\in\mathbb{L}_\kappa^{B\times T\times(D+1)}$, suitable for downstream tasks, where $\kappa_1$, $\kappa_2$ represent the curvatures before and after the transformation.

\subsection{Anomaly Detection}
\label{Anomaly Detection}

We estimate an anomaly score for each frame in a 3D point cloud video by measuring deviations from learned spatio-temporal patterns. Traditional Euclidean metrics, such as mean squared error or cosine similarity, often fail to capture the hierarchical geometry inherent in 3D data. To address this, we adopt the Lorentzian intrinsic distance in hyperbolic space, as defined in Equation~(\ref{eq19}), providing a more expressive and geometry-aware measure of deviation. This formulation naturally adapts to diverse scoring settings while preserving geometric consistency.
\begin{equation}
\label{eq19}
d_{\mathcal{L}}(\hat{\mathbf{X}}, \mathbf{P}_{T+1}) = \text{arcosh} \left( \kappa \langle \text{Decoder}(\hat{\mathbf{X}}), \mathbf{P}_{T+1} \rangle_{\mathcal{L}} \right)
\end{equation}
where $\text{Decoder}(\cdot)$ is the multilayer perceptron (MLP). In the classification setting, the decoder head directly outputs an anomaly score, and the Lorentzian intrinsic distance, defined in Equation~(\ref{eq19}), is employed as a hyperbolic loss to supervise the learning process.

\section{Experiments}

\subsection{Experiments Setup}

\begin{table*}
\captionsetup{justification=centering}
\centering
  \caption{
    Frame-level AUROC (\%) performance comparison of TIMo dataset. For each column, the top-performing method is marked in \textbf{bold}.}
  \label{tab:2}
    \begin{tabular}{c >{\centering\arraybackslash}m{5em} | >{\centering\arraybackslash}m{5em} >{\centering\arraybackslash}m{5em} >{\centering\arraybackslash}m{5em} | >{\centering\arraybackslash}m{2.5em} >{\centering\arraybackslash}m{2.5em}}
    \toprule
    \multirow{2}{*}{Method} & 
    \multirow{2}{*}{\makecell{Point Cloud\\Extractor}} & \multirow{2}{*}{\makecell{Aggressive\\Behavior}}&
    \multirow{2}{*}{\makecell{Medical\\Issue}} & 
    \multirow{2}{*}{\makecell{Left-Behind\\Objects}} & 
    \multicolumn{2}{c}{Total} \\
    \cline{6-7}
    & & & & & ES & HS\\
    \midrule
    CAE & -- & --& --& --& 66.4 & --\\
    ConvLSTM & -- & --& --& --& 62.8 & --\\
    R-CAE & -- & 76.8& 48.0& 66.6& 66.4 & --\\
    P-CAE & -- & 79.3& 59.9& 73.4& 71.4& --\\
    R-ViT-AE & -- & 68.3& 53.2& 71.8& 64.9& --\\
    P-ViT-AE & -- & 68.9& 53.6& 72.6& 65.1& --\\
    P-ConvLSTM & -- & 62.2& 50.9& 64.9& 62.8& --\\
    \midrule
    \multirow{3}{*}{\textbf{HyPCV-Former}} 
        & PointNet   &\heatcellA{80.4}{1} & \heatcellA{75.9}{1}& \heatcellA{77.5}{1}& \heatcellA{75.6}{1}&\heatcellA{77.3}{1}\\
        & PointMLP   & \heatcellA{79.7}{0}&\heatcellA{72.7}{0} & \heatcellA{75.4}{0}& \heatcellA{73.9}{0}&\heatcellA{74.9}{0}\\
        & DGCNN      & \heatcellA{80.0}{0}& \heatcellA{72.7}{0}& \heatcellA{76.1}{0}& \heatcellA{74.4}{0}&\heatcellA{75.5}{0}\\
    \bottomrule
    \end{tabular}
\end{table*}

\begin{table}
\captionsetup{justification=centering}
\centering
  \caption{
    Frame-level AUROC (\%) performance comparison of DAD dataset. For each column, the top-performing method is marked in \textbf{bold}.}
  \label{tab:22}
    \begin{tabular}{c >{\centering\arraybackslash}m{4.4em} | >{\centering\arraybackslash}m{2em} >{\centering\arraybackslash}m{2em}}
    \toprule
    \multirow{2}{*}{Method} & 
    \multirow{2}{*}{\makecell{Point Cloud\\Extractor}} & 
    \multicolumn{2}{c}{Total} \\
    \cline{3-4}
    & & ES & HS\\
    \midrule
    MobileNetV1 2.0× & -- & 90.18 & --\\
    MobileNetV2 1.0× & -- & 88.99 & --\\
    ShuffleNetV1 2.0× & -- & 88.69 & --\\
    ShuffleNetV2 2.0× & -- & 90.02& --\\
    ResNet-18 (from scratch) & -- & 89.96& --\\
    ResNet-18 (pre-trained) & -- & 90.20& --\\
    ResNet-18 (post-processed) & -- & 90.20& --\\
    \midrule
    \multirow{3}{*}{\textbf{HyPCV-Former}} 
        & PointNet   &\heatcellB{94.89}{1} & \heatcellB{95.55}{1}\\
        & PointMLP   &\heatcellB{88.60}{0} &\heatcellB{90.57}{0}\\
        & DGCNN      &\heatcellB{93.92}{0} & \heatcellB{95.08}{0}\\
    \bottomrule
    \end{tabular}
\end{table}

\subsubsection{Dataset} 
We evaluate our method on two public datasets designed for 3D video anomaly detection: TIMo~\cite{schneider2022timo} and DAD~\cite{kopuklu2021driver}. Currently, there exists no dedicated dataset for anomaly detection in 3D point cloud videos. To address this limitation, we convert depth video datasets into 3D point clouds via back-projection and evaluate our method accordingly. All baseline methods used for comparison are originally designed for depth video anomaly detection, as there is currently no publicly available method specifically developed for anomaly detection in 3D point cloud videos.

TIMo contains over 1500 sequences captured by an Azure Kinect depth camera, covering both normal and anomalous activities across tilted camera viewpoint. Invalid or missing depth values are removed. Following previous works~\cite{schneider2022unsupervised,he2024point}, we use 909 normal sequences for training and 679 sequences for testing, among which 569 are anomalous. 

DAD is a large-scale video dataset for monitoring anomalous driver behaviors in a simulated driving environment. In this work, we use only the front-view depth modality, which captures the driver’s upper body and head movements. The training set includes recordings from 25 subjects, and the test set comprises 36 sequences from 6 unseen subjects, where 16 types of anomalous actions not present in the training set occur unpredictably.

\begin{figure}[ht]
  \centering
  \includegraphics[width=\linewidth]{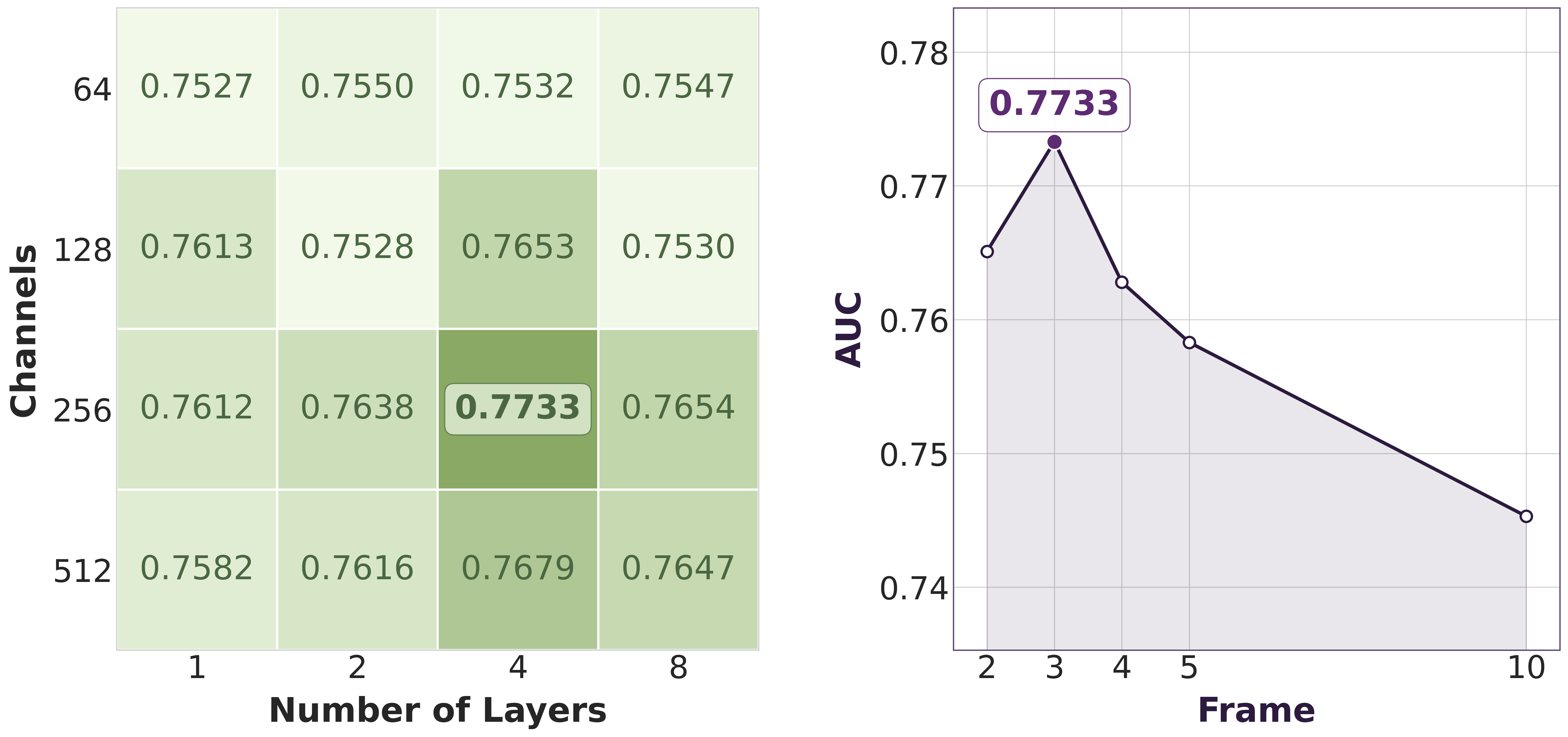}
  \caption{Parameter analysis of HyPCV-Former on TIMo dataset . Left: AUROC heatmap across different hyperbolic spatio-temporal transformer depths and channel widths. Right: Effect of varying the number of input frames on AUROC.}
  \label{fig:param_analysis}
\end{figure}

\begin{figure}[ht]
  \centering
  \includegraphics[width=\linewidth]{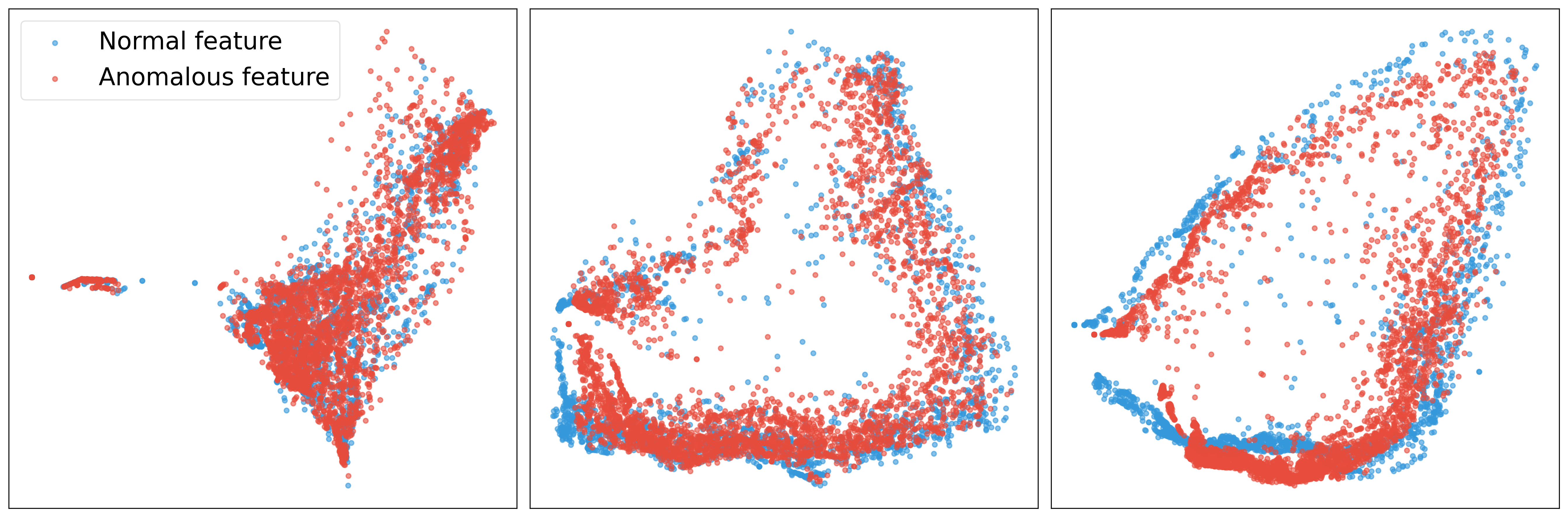}
  \caption{Isomap projection of feature distributions on TIMo dataset. Left: Raw point cloud features. Middle: Features trained with MSE in Euclidean space. Right: Features trained with hyperbolic geometry and Lorentzian distance.}
  \label{fig:feature_vis}
\end{figure}

\subsubsection{Implementation Details} HyPCV-Former is implemented in PyTorch and trained using distributed data parallel on 4 NVIDIA RTX 4090 GPUs with CUDA 12.2. Each input sequence consists of 3 consecutive frames, each downsampled to N = 2048 foreground points. A detailed explanation of this configuration is provided in the Appendix~\ref{A2}. The hyperbolic spatio-temporal transformation consists of 4 hyperbolic spatio-temporal transformer layers with 8 HMHA heads. The model is trained for 400 epochs using the AdamW optimizer with an initial learning rate of $1 \times 10^{-4}$ and cosine annealing schedule. We set variable curvatures as trainable parameters, following the design principle introduced in~\cite{yang2024hypformer}.

\subsubsection{Evaluation Metrics} We use the area under the ROC curve (AUROC) to evaluate the performance of HyPCV-Former following the previous methods in video anomaly detection~\cite{schneider2022timo,kopuklu2021driver}. Anomaly scores are computed based on the Lorentzian intrinsic distance in hyperbolic space. On the TIMo dataset, we adopt a prediction-based strategy and measure the distance between predicted and actual point clouds in the final frame. On the DAD dataset, we employ a classification-based approach, where the anomaly score is derived from the distance to normal class prototypes in the embedding space.

\begin{table*}
\captionsetup{justification=centering}
\centering
\caption{Ablation on the choice of space and loss function on TIMo dataset. The
top-performing choice is marked in \textbf{bold}.}
\label{tab:3}
\begin{tabular}{cc|cc|c|c|c}
\toprule
\multicolumn{2}{c|}{Space} & \multicolumn{2}{c|}{Loss Function}& \multirow{2}{*}{\makecell{\\Params }}& \multirow{2}{*}{\makecell{\\GFLOPS} } & \multirow{2}{*}{\makecell{\\HyPCV-Former\\{(\%)} }} \\
\cmidrule(lr){1-4}
Euclidean & Hyperbolic & MSE & \makecell{Lorentzian \\ Intrinsic Distance} & &  \\
\midrule
\checkmark &             & \checkmark &             &5.37M &1.534 & 75.4 \\
\checkmark &             &  &     \checkmark        &5.37M &1.534 & 77.0 \\
           & \checkmark  & \checkmark &             &5.92M &0.765 & 77.1 \\
           & \checkmark  &            & \checkmark  &5.92M & 0.765& \textbf{77.3} \\
\bottomrule
\end{tabular}
\end{table*}

\begin{figure*}[t]
    \centering
    \subfloat[Anomaly scores for sequence A0403]{
        \includegraphics[width=0.32\textwidth]{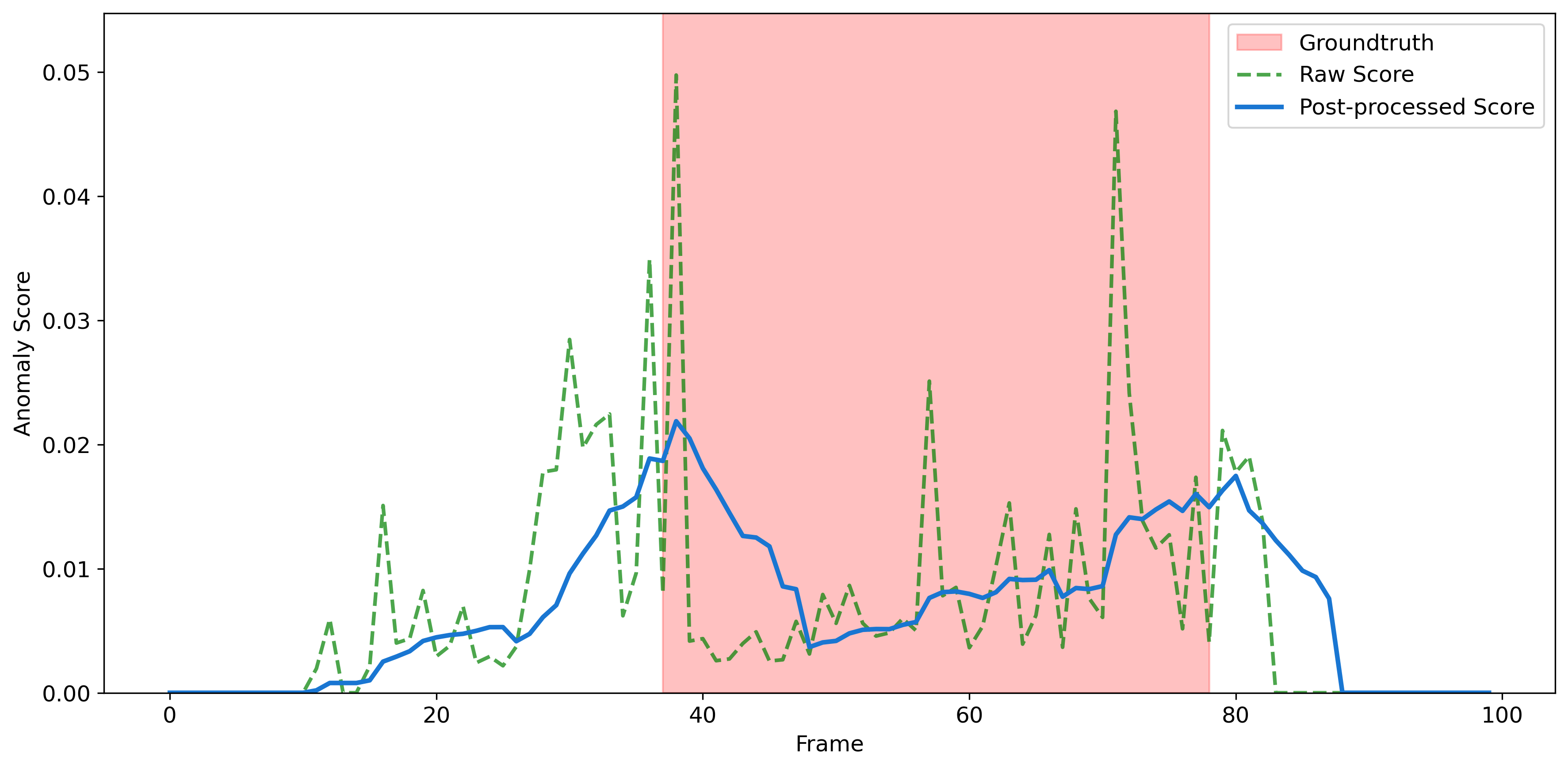}
    }
    \hfill
    \subfloat[Anomaly scores for sequence A0387]{
        \includegraphics[width=0.32\textwidth]{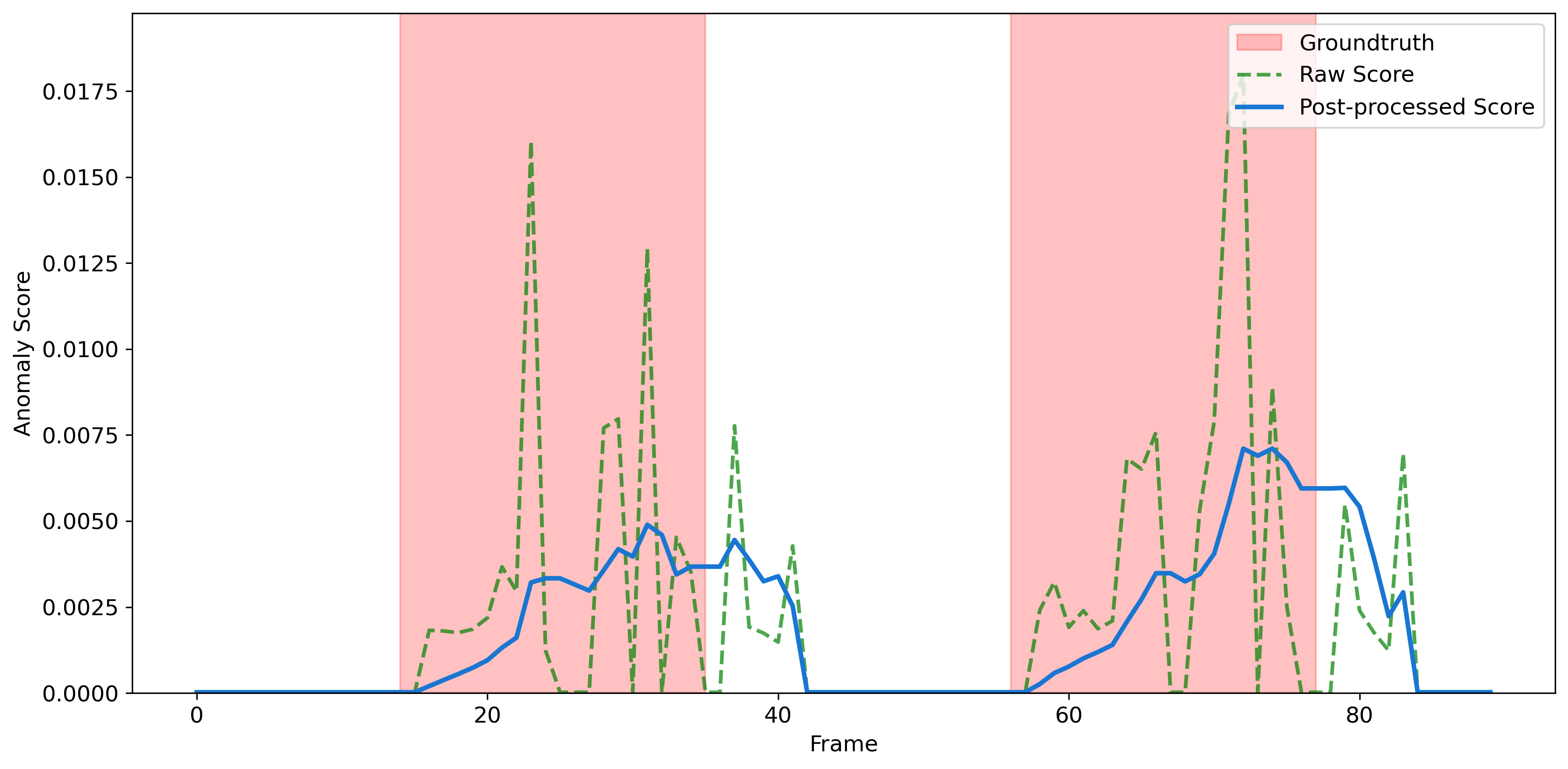}
    }
    \hfill
    \subfloat[Anomaly scores for sequence A0311]{
        \includegraphics[width=0.32\textwidth]{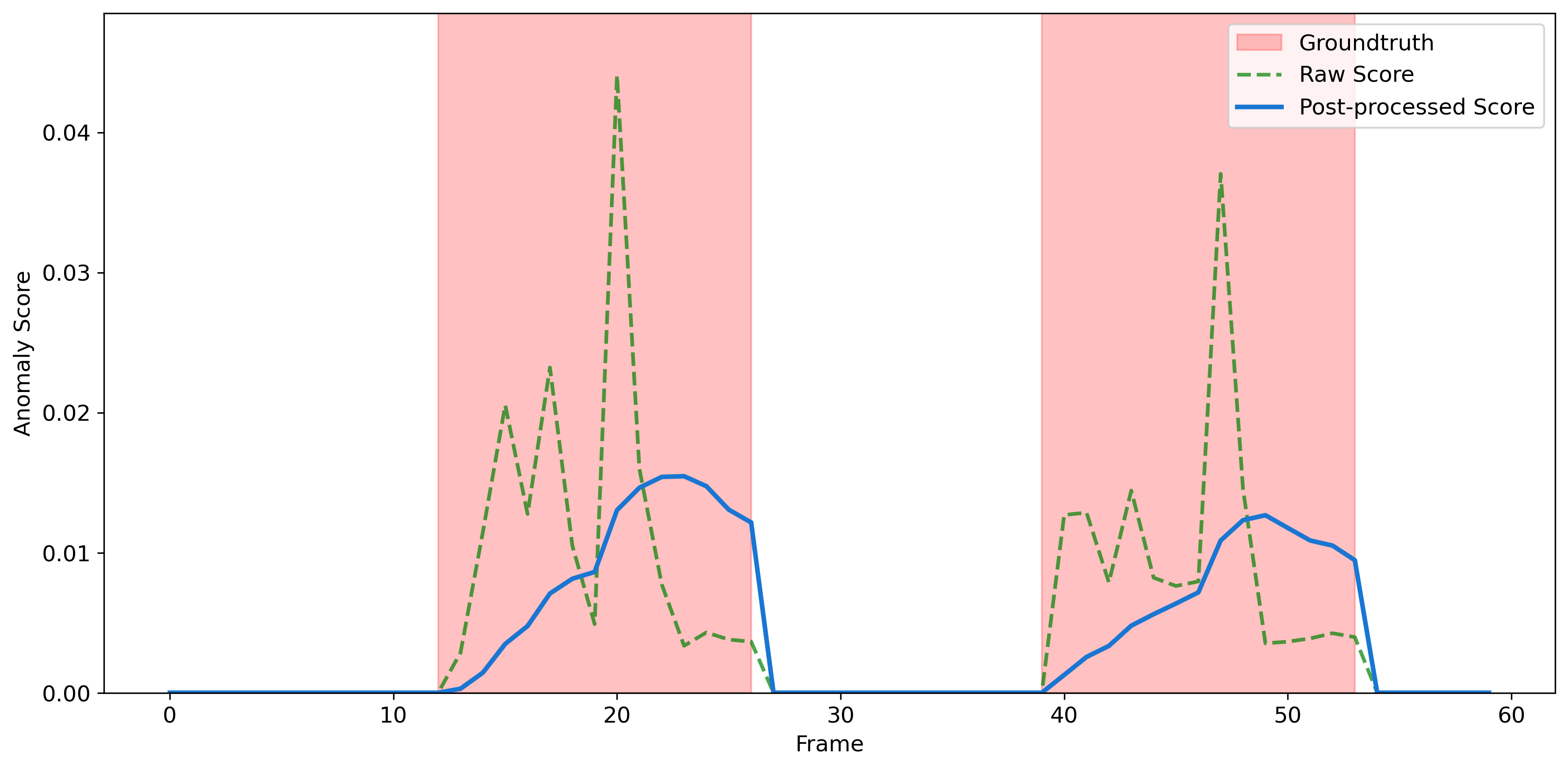}
    }
    \caption{
    Qualitative results for three anomaly categories in TIMo.
    (a) Anomaly scores for sequence A0403 (aggressive behavior), (b) Anomaly scores for sequence A0387 (medical issue), (c) Anomaly scores for sequence A0311 (left-behind object). 
    }
    \label{fig:5}
\end{figure*}

\begin{figure*}[t]
    \centering
    \subfloat[Anomaly scores for sequence val05-rec6]{
        \includegraphics[width=0.32\textwidth]{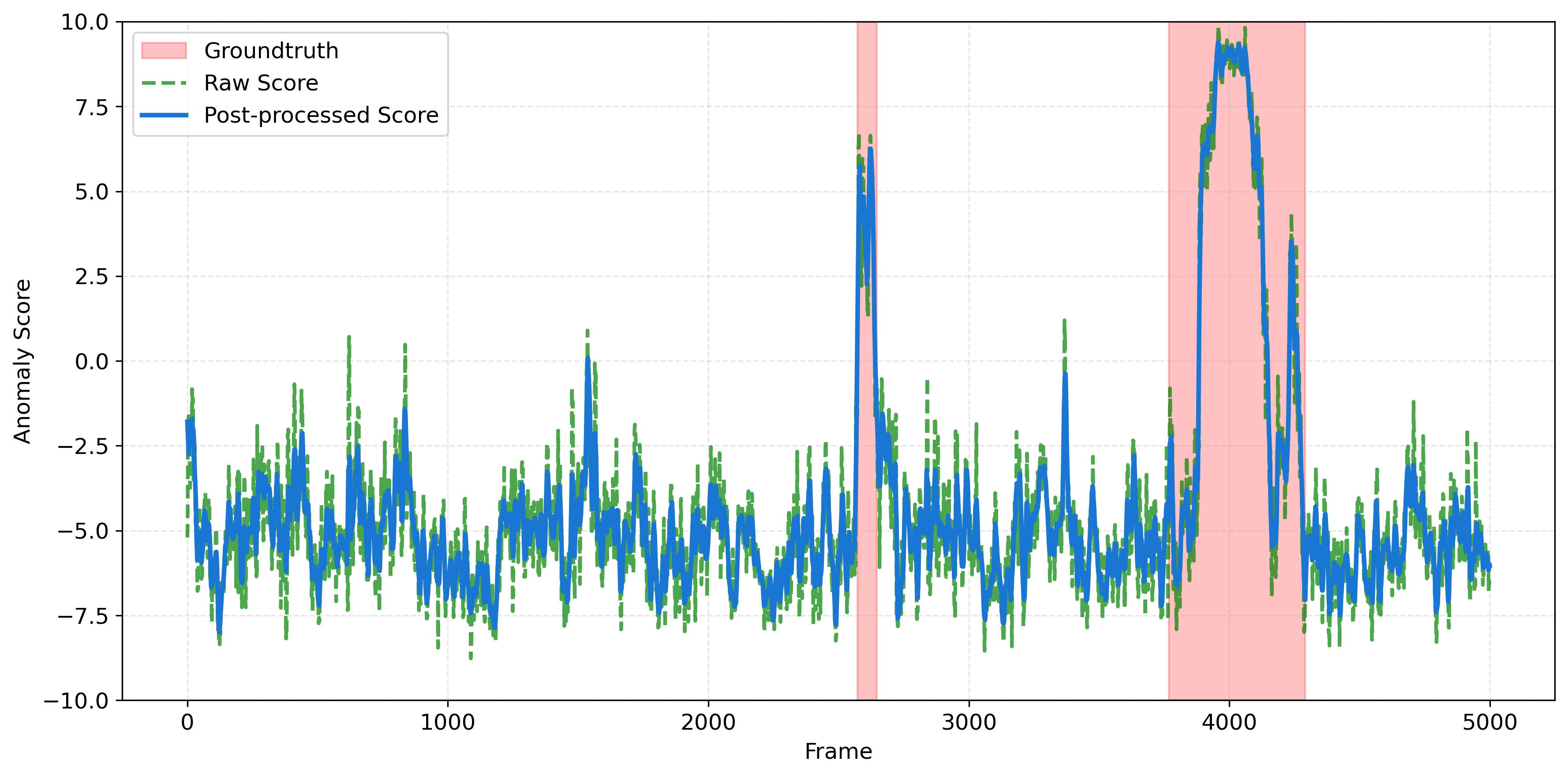}
    }
    \hfill
    \subfloat[Anomaly scores for sequence val05-rec3]{
        \includegraphics[width=0.32\textwidth]{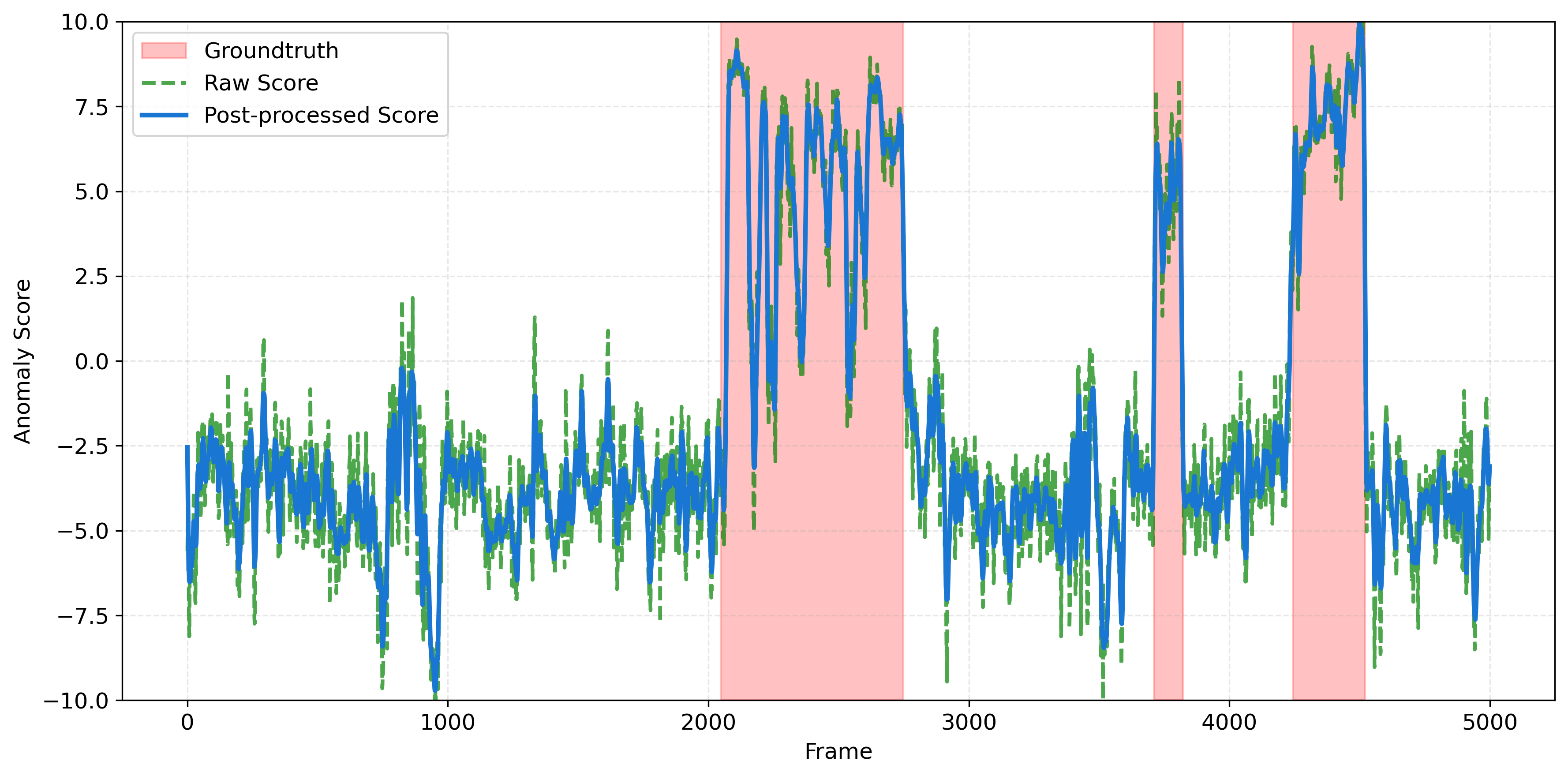}
    }
    \hfill
    \subfloat[Anomaly scores for sequence val04-rec6]{
        \includegraphics[width=0.32\textwidth]{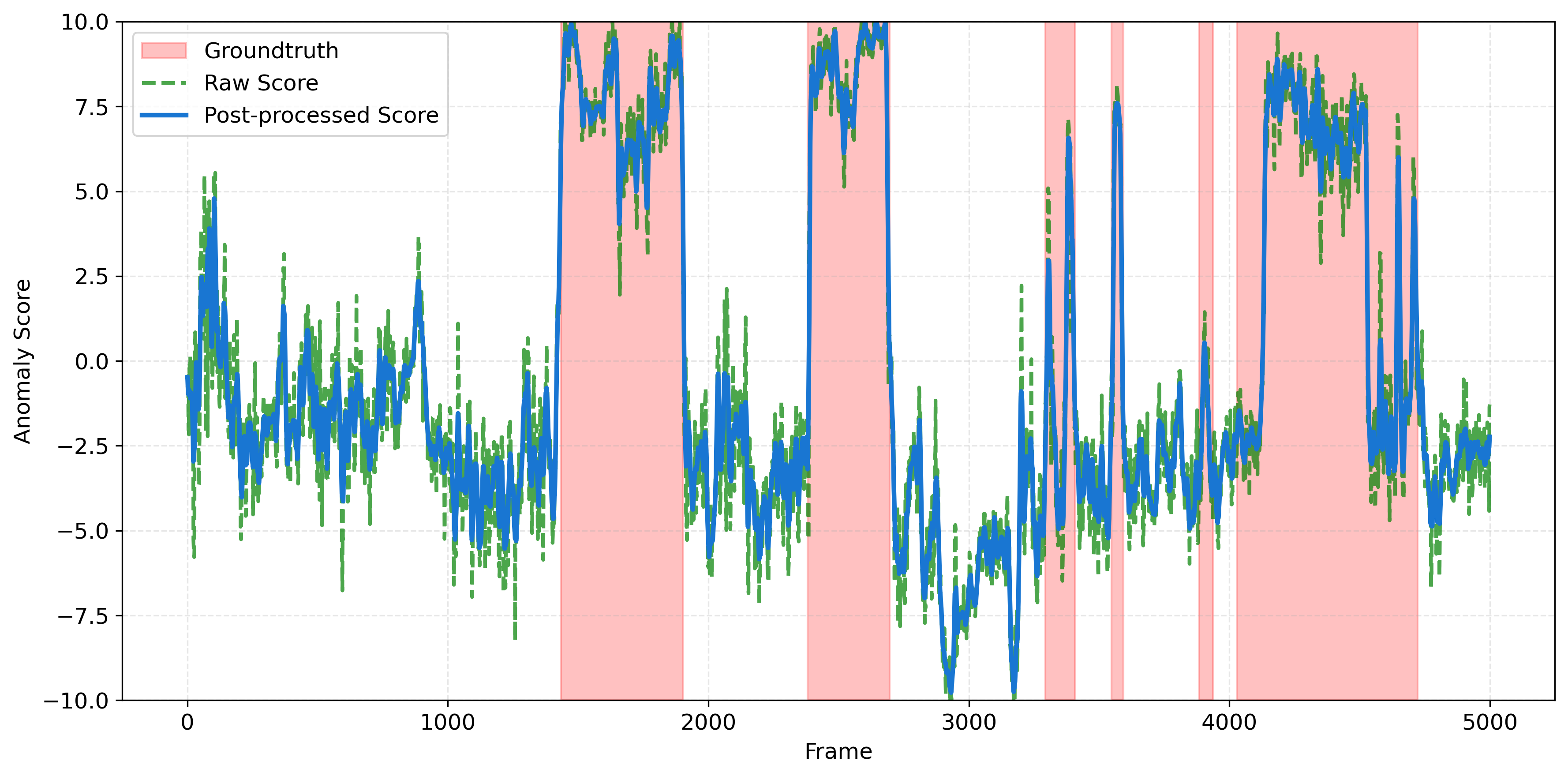}
    }
    \caption{
    Qualitative results for three anomaly records in DAD.
    (a) Video recording \#6 of subject \#5, (b) Video recording \#3 of subject \#5, (c) Video recording \#6 of subject \#4. 
    }
    \label{fig:55}
\end{figure*}

To reduce local noise and improve temporal consistency, we apply a simple post-processing step based on a moving average. Specifically, for each frame, the final score is computed as the average of the current and previous $w-1$ scores, where the window size $w$ is set to 10. This temporal smoothing helps stabilize predictions and better capture sustained abnormal behaviors across consecutive frames.

\subsection{Comparisons with State-of-the-art Methods}

To evaluate the effectiveness of our proposed HyPCV-Former, we compare it with several benchmarks on the TIMo dataset and DAD dataset, respectively. The introduction of benchmarks can be found in Appendix~\ref{A3}. As TIMo and DAD are originally depth video datasets, all baseline methods are implemented and evaluated on depth image sequences, as shown in Table~\ref{tab:2} and Table~\ref{tab:22}. To ensure consistency with prior work~\cite{schneider2022unsupervised}, we adopt the depth-based foreground mask generation method proposed by Braham \textit{et al.}~\cite{braham2014physically} in TIMo dataset. The detailed procedure is provided in Appendix~\ref{A4}. Following the previous work~\cite{schneider2022unsupervised}, we adopt the categorization of anomalies in TIMo into three types, including aggressive behavior, medical issue, and left-behind objects. We report results for each category as well as the total performance, where all anomaly types are evaluated together as a unified test set. Note that the total score is not an average, but reflects the overall detection ability of each method across diverse anomaly types.

As shown in Table~\ref{tab:2} and Table~\ref{tab:22}, HyPCV-Former consistently outperforms prior methods. When using PointNet as the point cloud feature extractor, our method achieves the best results in both category-wise and overall evaluations. We also assess performance using PointMLP and DGCNN, observing that PointNet offers more stable and discriminative representations within our framework. We further examine the effect of the distance metric used to compute prediction errors. Specifically, ES refers to Euclidean space distance using MSE, while HS corresponds to hyperbolic space distance computed using the Lorentzian intrinsic distance. The results show that hyperbolic distance yields consistently better performance, underscoring the benefit of modeling in curved geometry for 3D point cloud video anomaly detection.

It is also worth noting that we do not include the results of F-MSE and W-MSE losses proposed in~\cite{schneider2022unsupervised}, as these loss functions are specifically designed for depth images and do not generalize to point cloud representations. For a fair comparison, all baseline methods and our variants are evaluated using $L_2$-based loss.

\subsection{Additional Results and Analysis}

\subsubsection{Ablation Study} To evaluate the effectiveness of different components of HyPCV-Former, we conduct ablation studies focusing on the choice of geometric space and loss function. The results are summarized in Table~\ref{tab:3}. We first analyze the effect of using different geometric spaces. When applying only Euclidean space with MSE loss, the AUROC reaches $75.4\%$. Switching to hyperbolic space while keeping MSE as the loss function improves the performance to $77.0\%$, demonstrating the benefit of hyperbolic representation in capturing complex spatio-temporal structures. We then replace MSE with the Lorentzian intrinsic distance. This results in a further increase to $77.3\%$, showing that using a loss function aligned with the hyperbolic geometry leads to better anomaly discrimination. 

To validate the effectiveness of the key components in our proposed HMHA architecture, we conduct ablation studies on both the hyperbolic positional encoding and the HTC module, as shown in Table~\ref{tab:ablation}. The HyPCV-Former using our hyperbolic positional encoding achieves the highest AUC of $77.3\%$, outperforming both standard sinusoidal encoding and the variant without any positional encoding. The HTC-based HyPCV-Former consistently outperforms the standard variant, confirming that the curvature-aware transformation in hyperbolic space helps retain geometric consistency and provides a more expressive representation for spatiotemporal modeling.

\begin{table}[ht]
\centering
\caption{Ablation study on positional encoding and linear transformation on TIMo dataset. The top-performing choice is marked in \textbf{bold}.}
\begin{tabular}{c|c|c}
\toprule
\textbf{Module} & \textbf{Type} & \textbf{AUC (\%)} \\
\midrule
\multirow{3}{*}{\makecell{Positional \\ Encoding}} 
    & Hyperbolic-based & \textbf{77.3} \\
    & Standard-based & 77.0 \\
    & None & 76.4 \\
\midrule
\multirow{2}{*}{\makecell{Linear \\ Transformation}} 
    & HTC-based & \textbf{77.3} \\
    & Standard Linear-based & 76.8 \\
\bottomrule
\end{tabular}
\label{tab:ablation}
\end{table}

\subsubsection{Parameter Analysis} We further investigate the influence of several key hyperparameters on the performance of HyPCV-Former, including the number of hyperbolic spatio-temporal transformer layers, feature channels, and the input frame window size. As shown in the left subfigure of Figure~\ref{fig:param_analysis}, we conduct a grid search over different combinations of transformer layers and channel dimensions. We observe that the model achieves the best performance when using 4 layers and 256 channels, reaching an AUROC of 77.3\%. The right subfigure of Figure~\ref{fig:param_analysis} illustrates the impact of varying the number of input frames from 2 to 10. We find that the model performs best when using 3 frames for temporal modeling. While increasing the frame length initially improves performance due to more temporal context, excessive length introduces noise and redundancy, leading to a decline in accuracy.

\subsection{Visualization and Qualitative Analysis} 

\subsubsection{Feature Discrimination Visualization} To further demonstrate the discriminative capability of the learned features under different training settings, we visualize the representations of normal and anomalous frames using Isomap projection. Figure~\ref{fig:feature_vis} illustrates the feature distributions under three settings: (1) direct output of the point cloud extractor, (2) features learned with MSE loss in the Euclidean space, and (3) features learned with Lorentzian intrinsic distance in the hyperbolic space.

As shown in the left plot, when using the point cloud encoder alone, the extracted features for normal and anomalous samples show significant overlap, indicating limited separation in the latent space. Introducing MSE loss slightly improves the feature separability, yet the two classes still exhibit considerable mixing. In contrast, our method—fully operating in hyperbolic space with geometry-consistent distance—yields a much clearer separation. The boundary between normal and abnormal features becomes more distinct, validating that the hierarchical structures of temporal dynamics are better captured when learning is performed intrinsically in hyperbolic geometry.

\subsubsection{Qualitative Visualizations} We visualize anomaly scores from three representative videos, each from a different anomaly category in the TIMo dataset. Figure~\ref{fig:5} (a) shows an aggressive behavior case, (b) is a medical issue, and (c) is luggage left behind. The qualitative results of three different video recordings in DAD dataset are visualized in Figure~\ref{fig:55}. In each plot, the green dashed line is the raw anomaly score, and the blue solid line is the post-processed score. The red shaded areas represent ground truth anomalies. Post-processing clearly reduces noise and improves temporal consistency.

\section{Conclusions}

In this paper, we propose HyPCV-Former, a novel hyperbolic spatio-temporal transformer-based framework for anomaly detection in 3D point cloud videos. Our method leverages the rich spatio-temporal structure of point cloud sequences by first extracting per-frame features using point-based networks, and subsequently projecting them into a Lorentzian hyperbolic space. We introduce an HMHA mechanism that effectively captures temporal dependencies within the hyperbolic manifold, enabling more discriminative representations for anomaly prediction. Extensive experiments show that HyPCV-Former achieves state-of-the-art performance across multiple anomaly categories. In addition, ablation studies and qualitative visualizations further verify the advantages of hyperbolic modeling, the choice of loss functions, and our spatial-temporal design.

In future work, we plan to explore adaptive curvature learning to further enhance the geometric expressiveness of our model. We also aim to extend our framework to multi-view or multi-modal 3D video understanding tasks, such as action recognition and behavior forecasting.

\bibliography{aaai2026}

\appendix

\section{Formulations of HTC and HRC}
\label{A1}

HTC enables curvature-aware linear projections directly within the Lorentz model, facilitating consistent mappings across hyperbolic spaces with varying curvature. HRC redefines key non-linear operations—such as activation, normalization, and dropout—by acting only on the spatial feature dimensions, preserving the causal structure of Lorentzian geometry. These modules jointly provide a principled foundation for learning spatio-temporal representations fully in hyperbolic space.

The hyperbolic transformation with curvatures (HTC) module for hyperbolic embeddings is defined as follows~\cite{yang2024hypformer}:

\begin{equation}
    \text{HTC}(\mathbf{x}^{\mathbb{L}}_{t};\mathbf{W}_h,\kappa_1,\kappa_2)=\left(\sqrt{\frac{\kappa_1}{\kappa_2}\|\mathbf{x}^{\mathbb{L}}_{t}\mathbf{W}_h\|_2^2-\frac{1}{\kappa_2}},\,\sqrt{\frac{\kappa_1}{\kappa_2}}\mathbf{x}^{\mathbb{L}}_{t}\mathbf{W}_h\right),
\end{equation}
where $\mathbf{W}_h$ denotes the learnable parameters for each attention head, and $\kappa_1$, $\kappa_2$ represent the curvatures before and  after the transformation.

The HRC ensures stable and effective manifold embeddings through several hyperbolic-specific operations, including hyperbolic normalization, activation, and dropout. Specifically, the HRC module is defined as follows:
\begin{equation}
\text{HRC}(\mathbf{O};f_r,\kappa_1,\kappa_2)=\left(\sqrt{\frac{\kappa_1}{\kappa_2}\|f_r(\mathbf{O})\|_2^2-\frac{1}{\kappa_2}},\,\sqrt{\frac{\kappa_1}{\kappa_2}}f_r(\mathbf{O})\right),
\end{equation}

\section{Justification for Choosing 2048 Foreground Points}
\label{A2}

Table~\ref{tab:point_ablation} presents a quantitative comparison of model performance and computational complexity across different numbers of input points per frame. As shown, increasing the number of points from 512 to 8192 results in a steady improvement in AUC, indicating enhanced anomaly detection capability. However, this performance gain comes at the cost of significantly increased model size and floating point operations (FLOP). For instance, moving from 2048 to 8192 points nearly doubles the AUC gain by only 0.8 points, while FLOPs more than quadruple.

\begin{table}[ht]
\centering
\caption{Model performance and computational cost under different point numbers.}
\begin{tabular}{c|c|c|c}
\toprule
Points num & AUC (\%) & Params & FLOP \\
\midrule
512   & 74.2 & 4.741M  & 197.832M \\
1024  & 75.7 & 5.136M  & 386.576M \\
2048  & 77.3 & 5.925M  & 765.636M \\
4096  & 77.8 & 7.504M & 1.524G \\
8192  & 78.1 & 10.662M & 3.040G \\
\bottomrule
\end{tabular}
\label{tab:point_ablation}
\end{table}

To strike a balance between performance and computational cost, we select 2048 points per frame. This setting provides a favorable trade-off and aligns with previous method~\cite{he2024point}, ensuring consistency and comparability in experimental evaluation.

\section{Overview of Baseline Methods}
\label{A3}

To facilitate fair comparison, we adopt a series of baseline models originally introduced in Schneider \textit{et al.}~\cite{schneider2022unsupervised} for unsupervised video anomaly detection on depth data. Below, we briefly describe each method:

\begin{itemize}

\item {\textbf{CAE (Convolutional Autoencoder)}. This model reconstructs individual frames using a symmetric convolutional encoder-decoder architecture. The anomaly score is computed based on the reconstruction error, under the assumption that anomalies cannot be accurately reconstructed due to their deviation from learned normal patterns.} 

\item {\textbf{ConvLSTM}. Based on the ConvLSTM framework~\cite{shi2015convolutional}, this model incorporates temporal dependencies by leveraging convolutional operations within LSTM cells. It processes a sequence of frames and predicts the next frame, using the prediction error as the anomaly score.}

\item {\textbf{R-CAE (Reconstruction CAE)}. A specific implementation of CAE tailored for depth frames, R-CAE aims to reconstruct a single input frame. It serves as a representative model for frame-wise spatial reconstruction in anomaly detection tasks.}

\item {\textbf{P-CAE (Prediction CAE)}. P-CAE extends the CAE paradigm to a prediction setting. It takes a short sequence of preceding frames as input and predicts the subsequent frame. The divergence between the predicted and actual frame is used to detect anomalous behavior.}

\item {\textbf{R-ViT-AE (Reconstruction Vision Transformer Autoencoder)}. Utilizing the Vision Transformer architecture~\cite{dosovitskiy2020image}, R-ViT-AE reconstructs the current frame using self-attention mechanisms over image patches. It explores the potential of transformer-based architectures for spatial feature modeling in depth video.}

\item {\textbf{P-ViT-AE (Prediction Vision Transformer Autoencoder)}. Similar in architecture to R-ViT-AE, this model predicts the next frame based on a sequence of prior frames. It aims to model both spatial and short-term temporal dynamics using transformer layers.}

\item {\textbf{P-ConvLSTM (Prediction ConvLSTM)}. This model combines the strengths of convolution and LSTM to capture spatiotemporal dependencies across frames. It is designed to predict the next frame from a given sequence and is particularly suitable for depth video data where motion continuity is a key anomaly cue.}
\end{itemize}

To provide a comprehensive evaluation, we compare our method with several baseline models originally reported in the DAD benchmark~\cite{kopuklu2021driver}, including both lightweight and standard architectures:
\begin{itemize}
\item {\textbf{MobileNetV1 2.0× and MobileNetV2 1.0×} are compact CNNs designed for efficient inference on resource-constrained devices~\cite{sandler2018mobilenetv2}.}
\item {\textbf{ShuffleNetV1 2.0× and ShuffleNetV2 2.0×} are lightweight architectures that utilize group convolutions and channel shuffling to reduce computational complexity~\cite{zhang2018shufflenet}.}
\item {\textbf{ResNet-18 (from scratch)} is a standard residual network trained directly on the DAD dataset without pre-initialization~\cite{ou2019simulation}. It serves as a baseline for evaluating the effectiveness of deeper architectures under limited supervision.}
\item {\textbf{ResNet-18 (pre-trained)} is initialized with ImageNet-pretrained weights and then fine-tuned on the DAD dataset~\cite{ou2019simulation}. This variant leverages transfer learning to boost performance.}
\item {\textbf{ResNet-18 (post-processed)} is based on the pre-trained ResNet-18 but incorporates a simple temporal smoothing technique applied to frame-wise predictions to reduce noise and improve anomaly detection consistency~\cite{ou2019simulation}.}
\end{itemize}

\section{Foreground Mask Generation}
\label{A4}

Foreground segmentation is a critical step in video anomaly detection because anomalous activities typically occur within the foreground region of the observed scene. Traditional methods designed for RGB images often struggle with background segmentation, especially under dynamic environmental conditions. Thus, utilizing depth information significantly simplifies and improves the reliability of foreground segmentation~\cite{schneider2022unsupervised,he2024point}.

In our study, we adopt the depth-based pixel-wise background subtraction method proposed by Braham \textit{et al.}~\cite{braham2014physically}. This physically motivated model leverages the inherent properties of range images, such as their robustness to illumination changes and distinct foreground-background depth discontinuities, to achieve reliable segmentation. The parameters we used are given in Table~\ref{tab:1}.

\begin{table}[ht]
  \centering
  \caption{Parameter settings for the foreground segmentation method following Braham \textit{et al.}~\cite{braham2014physically}}
  \label{tab:1}
  \begin{tabular}{c|c|c|c|c|c|c}
    \hline
    Param. & $K$ & $K_{\text{kinect}}$ & $\Delta P_{\text{max}}$ & $\alpha$ & $T_{\text{W}}$ & $N_{\text{H}}$ \\ \hline
    Value & $1.25$ & $5\cdot10^{-4}$ & $100$ & $0.4$ & $300$ & $90$ \\ \hline
  \end{tabular}
\end{table}

\end{document}